\documentclass[conference]{IEEEtran} \IEEEoverridecommandlockouts

\usepackage{cite} \usepackage{amsmath,amssymb,amsfonts} \usepackage{algorithmic} \usepackage{graphicx} \graphicspath{{figures/}} \usepackage{textcomp} \usepackage{xcolor} \usepackage{booktabs} \usepackage{multirow} \usepackage{diagbox} \usepackage{url} \usepackage{algorithm} \def\BibTeX{{\rm B\kern-.05em{\sc i\kern-.025em b}\kern-.08em T\kern-.1667em\lower.7ex\hbox{E}\kern-.125emX}}

\begin{document}

\title{NoRIN: Backbone-Adaptive Reversible Normalization for Time-Series Forecasting}

\author{%
  \IEEEauthorblockN{%
    Shun Zhang\IEEEauthorrefmark{1}\IEEEauthorrefmark{2},
    Yuyang Xiao\IEEEauthorrefmark{1}\IEEEauthorrefmark{2},
  }
  \IEEEauthorblockA{%
    \IEEEauthorrefmark{1}\textit{Graduate School of China Academy of Engineering Physics}, Beijing 100193, China\\
    \IEEEauthorrefmark{2}\textit{Institute of Applied Physics and Computational Mathematics}, Beijing 100094, China\\
    }%
}

\maketitle

\begin{abstract}
Reversible instance normalization (RevIN) and its successors
(Dish-TS, SAN, FAN) have become the de facto plug-in for
time-series forecasting, yet the map they apply to each data point is
strictly affine, $x\mapsto ax+b$, so they cannot reshape the
underlying distribution---heavy tails remain heavy and skewness
remains uncorrected. We propose \textbf{NoRIN}, a non-linear
reversible normalization based on the
$\operatorname{arcsinh}$-form Johnson $S_U$ transform with two shape
parameters $(\delta,\varepsilon)$ that control tailedness and
skewness; the linear $Z$-score used by RevIN is recovered only in
the limit $\delta\!\to\!\infty$. Training $(\delta,\varepsilon)$
jointly with the backbone via gradient descent reliably pushes them
toward this linear limit within a few epochs---a phenomenon we name
the \emph{degeneration problem}: the forecasting loss is locally
indifferent to shape, and the high-capacity backbone compensates for
any monotone reparameterization of its input. NoRIN escapes the
degeneration by \emph{decoupling} shape selection from gradient
training: $(\delta,\varepsilon)$ are initialized by a closed-form
Slifker--Shapiro quantile fit and refined by Bayesian optimization on
the validation objective, while the inner training loop is identical
to standard RevIN-style training. Across six representative backbones $\times$ five
real-world datasets $\times$ three prediction horizons
($90$ configurations, $1{,}620$ training runs), NoRIN
significantly outperforms every linear-normalization
baseline (RevIN, SAN, Dish-TS, DeStat) at $p < 10^{-14}$
(paired Wilcoxon); the recovered $(\delta^\star,\varepsilon^\star)$
sit systematically far from the linear limit and vary in
a backbone-dependent way, supporting the thesis that
different backbones genuinely require different
normalization parameters to reach their best performance.
\end{abstract}


\section{Introduction}
\label{sec:intro}

Deep time-series forecasting has advanced rapidly in recent years, with Transformer- and MLP-based backbones such as PatchTST~\cite{nie2023patchtst}, iTransformer~\cite{liu2024itransformer}, and DLinear~\cite{zeng2023dlinear} repeatedly setting the state of the art on long-horizon benchmarks. A recurring observation across this literature is that \emph{how} the raw series enters the network is at least as consequential as the network itself. Real-world time series are pervasively non-stationary: mean, variance, and higher-order moments drift over time, so the distribution of a lookback window can differ substantially from that of the horizon it is meant to predict~\cite{kim2022revin,fan2023dishts,hu2023boosting}.

Reversible instance normalization (RevIN)~\cite{kim2022revin} has emerged as the de facto response to this challenge: each lookback window is standardized by its own mean and standard deviation before the forecaster, and the operation is inverted at the output. A stream of follow-up work has refined this recipe---Dish-TS~\cite{fan2023dishts} couples input- and output-space distributions, SAN~\cite{liu2023san} normalizes at the slice level, and FAN~\cite{ye2024fan} moves the operation into the frequency domain---all consistently improving forecasting accuracy across backbones.

However, these methods share an overlooked property: the per-element map always belongs to the \emph{affine} family $x \mapsto ax+b$. Even RevIN's optional learnable affine post-layer $(\gamma,\beta)$ leaves the composed transformation affine; the variants differ only in how $(a,b)$ are computed---per-instance, per-slice, or in the frequency domain---while the transformation \emph{family} itself is fixed and 
identical across backbones. A recent analysis further shows that standard RevIN can inflate MSE by as much as 683\% on outlier-rich datasets~\cite{huang2025noiseorsignal}.

The natural way to break this ceiling is to replace the affine normalizer with a \emph{non-linear, monotone, invertible} transformation: functions such as $\operatorname{arcsinh}$ behave linearly near zero but compress logarithmically for large values, so they can genuinely reshape the distribution---tame heavy tails, correct skewness---while preserving invertibility. Yet in the five years since RevIN, no genuinely non-linear reversible instance normalizer has become standard practice. The reason, we find, is not that the idea has been overlooked: it is that \emph{non-linear shape parameters cannot be learned end-to-end with the backbone}. Our own first attempt exposed the issue: when the shape parameters $(\delta,\varepsilon)$ of an $\operatorname{arcsinh}$-style transform are treated as learnable scalars trained jointly with the backbone, gradients systematically push them toward the linear limit, silently collapsing the non-linear normalizer back into RevIN---and delivering no benefit on precisely the heavy-tailed datasets where non-linearity should matter most. We call this the \emph{degeneration problem}: the forecasting loss is locally indifferent to shape, while the high-capacity backbone can compensate for any monotone reparameterization of its input by adjusting internal weights, so gradients are dominated by scale terms and drift steadily toward the easy-to-optimize Gaussian corner. This explains why the seemingly obvious idea of non-linear normalization has remained absent from deep forecasting.

We propose \textbf{NoRIN}---a non-linear reversible normalization based on the $\operatorname{arcsinh}$-form Johnson~$S_U$ transform~\cite{johnson1949systems} with two shape parameters $(\delta,\varepsilon)$ controlling tailedness and skewness. The linear $Z$-score used by RevIN is recovered only in the limit $\delta\!\to\!\infty$; for any finite $(\delta,\varepsilon)$ the map is genuinely non-linear and therefore capable of reshaping the distribution. Rather than imposing a single fixed transformation across all backbones, NoRIN lets each backbone use its own $(\delta,\varepsilon)$, selected by an outer process. To avoid the degeneration problem, we decouple parameter selection from gradient training: $(\delta,\varepsilon)$ is initialized by a closed-form Slifker--Shapiro quantile fit~\cite{slifker1980} and refined by Bayesian optimization~\cite{bergstra2011tpe,akiba2019optuna} against the true validation objective, while the inner training loop is identical to standard RevIN-style training. The framework is agnostic to both the specific non-linear family and the specific outer solver---JSU could be replaced, BO could be omitted---what matters is removing $(\delta,\varepsilon)$ from the gradient and exposing them to a search.

We further observe that the $(\delta^\star,\varepsilon^\star)$ values recovered by Bayesian optimization vary systematically across backbones (Sec.~\ref{sec:exp-main}). Backbones differ markedly in how much non-linear tail compression they prefer: attention-based backbones such as Informer~\cite{zhou2021informer} settle at substantially larger $\delta^\star$, while PatchTST~\cite{nie2023patchtst} and the linear baseline DLinear~\cite{zeng2023dlinear} converge near the lower search boundary at small $\delta^\star$, with iTransformer~\cite{liu2024itransformer} in between. This empirical pattern confirms that different backbones genuinely require different normalization parameters.

\paragraph*{Contributions.}
\begin{itemize}

  \item \textbf{Degeneration problem.} We discover and diagnose why
        naive non-linear extensions fail: end-to-end training
        systematically collapses non-linear shape parameters back to
        the linear regime, because the loss is locally indifferent to
        shape and the high-capacity backbone absorbs any monotone
        reparameterization. This explains the absence of non-linear
        reversible normalization in deep forecasting despite its
        apparent obviousness.
  \item \textbf{Non-linear normalization with decoupled shape
        optimization.} We replace RevIN's affine map with the
        $\operatorname{arcsinh}$-form Johnson~$S_U$ transform, governed
        by two shape parameters $(\delta,\varepsilon)$; the linear
        $Z$-score is recovered only in the limit $\delta\!\to\!\infty$,
        and at any finite $(\delta,\varepsilon)$ the transform is
        genuinely non-linear. To avoid degeneration,
        $(\delta,\varepsilon)$ is selected by an outer
        process---closed-form Slifker--Shapiro warm initialization
        followed by Bayesian optimization against the validation
        objective---while the inner training loop is unchanged. The
        resulting normalizer is drop-in compatible with any
        RevIN-using backbone.
  \item \textbf{Empirical validation across backbones and datasets.}
        Across six representative backbones $\times$ five real-world
        datasets $\times$ three prediction horizons (90 configurations,
        1,620 training runs), NoRIN significantly outperforms every
        linear-normalization baseline at $p<10^{-14}$ (paired
        Wilcoxon). The HPO-recovered $(\delta^\star,\varepsilon^\star)$
        values lie systematically far from the linear limit and vary
        in a backbone-dependent way, confirming that different
        backbones genuinely require different normalization parameters
        to reach their best performance---a regime that end-to-end
        gradient training cannot access.
\end{itemize}

\section{Related Work}
\label{sec:related}

\subsection{Reversible Instance Normalization and Its Linear Family}
Instance-wise normalization for time-series forecasting was popularized
by RevIN~\cite{kim2022revin}: each lookback window is standardized by
its own mean and standard deviation, the operation is reversed at the
horizon, and a small learnable affine layer provides minor post-hoc
adjustment. This simple recipe has become a near-universal component of
modern forecasting backbones, including PatchTST~\cite{nie2023patchtst},
iTransformer~\cite{liu2024itransformer}, and linear
baselines~\cite{zeng2023dlinear}.

Several works extend RevIN to address limitations of the single-mean/single-variance assumption. Non-stationary Transformer~\cite{liu2022nonstationary} stationarizes inputs while re-introducing learned non-stationary signals through attention. Dish-TS~\cite{fan2023dishts} frames the problem as a distribution shift between input and output space and learns two coefficient networks to describe each. SAN~\cite{liu2023san} argues that whole-instance statistics are too coarse and normalizes at the temporal-slice level while explicitly predicting future slice statistics. FAN~\cite{ye2024fan} moves the operation into the frequency domain and adaptively normalizes dominant frequency components to handle evolving seasonality. A concurrent analysis~\cite{huang2025noiseorsignal} studies RevIN's behavior under outliers and shows that replacing non-robust statistics with robust counterparts (e.g., median, MAD) avoids catastrophic failure on heavy-tailed data.

We emphasize a common thread across all of these methods: while they enrich \emph{where} and \emph{how often} statistics are computed, the transformation they apply to each data point remains \emph{affine} ($x \mapsto ax+b$). No matter how sophisticated the statistic estimator, an affine map cannot reshape a distribution---this is the ceiling shared by the entire RevIN family. NoRIN is \emph{orthogonal and complementary} to all of the above: it changes the \emph{functional form} of the normalizer from affine to non-linear-monotone, making it capable of reshaping the distribution itself, and can in principle be combined with any of the above granularity choices.

\subsection{Non-Linear Distribution Transforms}
Outside the deep-learning literature, non-linear variance-stabilizing
transforms have a long history. Box--Cox~\cite{box1964transformations} and Yeo--Johnson~\cite{yeo2000transformation} use a single
power parameter to correct skewness; the Johnson
system~\cite{johnson1949systems} extends this further by covering the
full $(\mathrm{skewness},\mathrm{kurtosis})$ plane via three disjoint
families ($S_U$ for unbounded, $S_B$ for bounded, $S_L$ for lognormal
limits). The Johnson~$S_U$ variant in particular expresses any
$S_U$-distributed variable as an
$\operatorname{arcsinh}$-reparameterized normal, making it naturally
invertible and differentiable---which is precisely why NoRIN adopts it
as the non-linear instantiation of choice. While Johnson transforms
are standard in statistical quality control and financial risk
modeling, to our knowledge they have not been used as reversible
instance normalizers in deep forecasting. This absence is not
accidental: the degeneration problem blocks direct end-to-end use, and
overcoming it is precisely the contribution of this paper. The
closest work in spirit is DeepAR-style probabilistic forecasters that
place heavy-tailed output distributions (e.g., Student-$t$) on the
prediction, but those act on the \emph{output} likelihood, not on the
\emph{input} representation.

\subsection{Decoupled Hyperparameter Optimization}
Bayesian optimization is a well-established framework for tuning
expensive black-box functions; the Tree-structured Parzen
Estimator (TPE)~\cite{bergstra2011tpe} and Gaussian-process-based
samplers excel in low-dimensional search spaces, and
Optuna~\cite{akiba2019optuna} provides a define-by-run interface
together with pruning and distributed execution, making it a de-facto
standard in the time-series community. In NoRIN, we treat the shape
parameters $(\delta,\varepsilon)$ as \emph{hyperparameters} of the
training procedure: they are frozen during inner training and selected
by an outer BO loop against validation MSE, warm-started from a
closed-form Slifker--Shapiro fit~\cite{slifker1980} that keeps the
search in a statistically sensible region. Our contribution is not BO
itself but the recognition that shape parameters \emph{should be
treated as hyperparameters}---a paradigm shift that is the key to
escaping the degeneration problem.

\section{Method}
\label{sec:method}

\begin{figure*}[!t]
  \centering
  \includegraphics[width=0.96\textwidth]{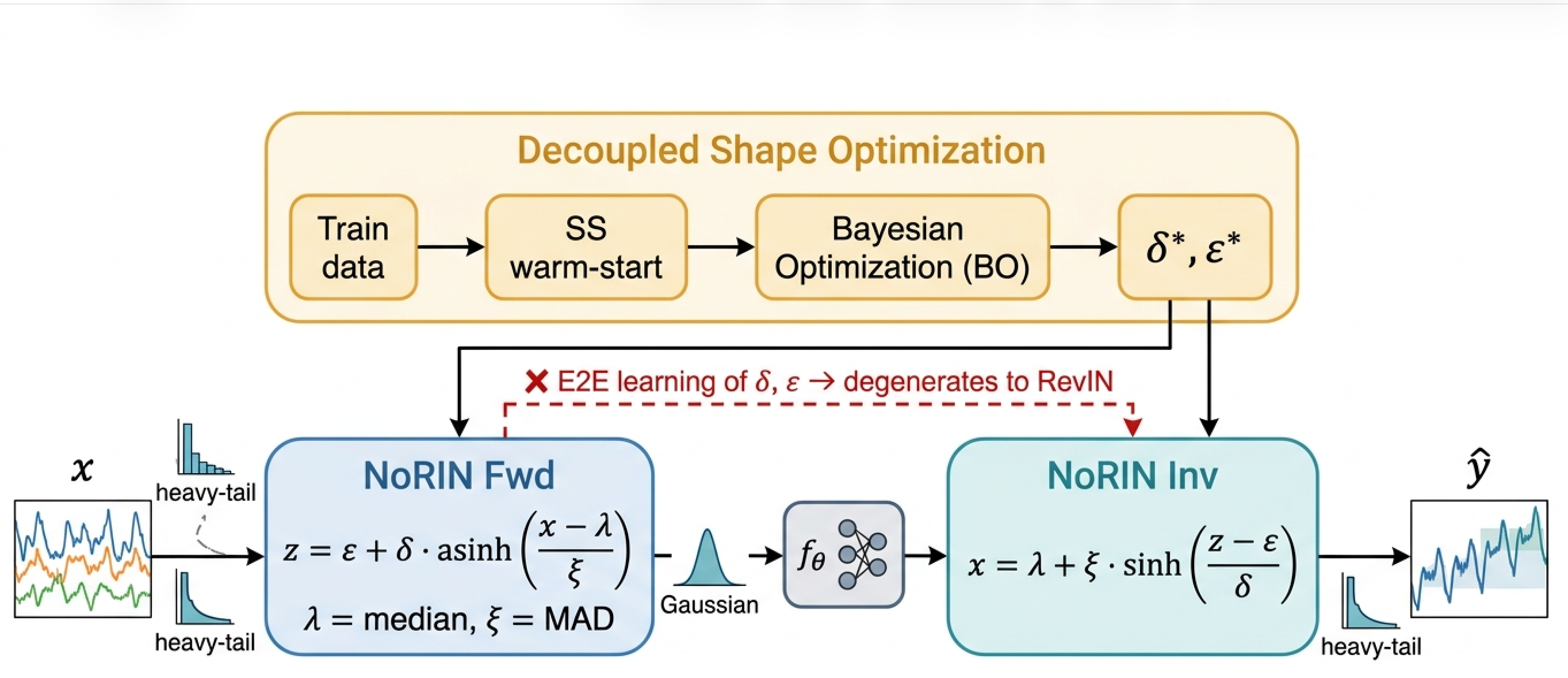}
  \caption{\textbf{NoRIN architecture overview.} Decoupled Shape
  Optimization (top) recovers the JSU shape parameters
  $(\delta^\star,\varepsilon^\star)$ once via Slifker--Shapiro
  warm-start followed by Bayesian optimization on validation MSE;
  these parameters are then \emph{frozen} and injected into both
  the forward (NoRIN Fwd) and inverse (NoRIN Inv) JSU transforms,
  which sandwich an arbitrary forecasting backbone $f_\theta$.
  Inputs $x$ are heavy-tailed; the JSU non-linearity reshapes them
  to approximately Gaussian for the backbone, and the inverse maps
  predictions back to the original heavy-tailed scale. End-to-end
  gradient learning of $(\delta,\varepsilon)$ (red dashed arrow)
  degenerates to RevIN (Sec.~\ref{sec:method-degen}); affine RevIN
  alone leaves the distribution shape unchanged.}
  \label{fig:architecture}
\end{figure*}

We first recall the general template of reversible instance normalization and emphasize that it does \emph{not} require linearity (Sec.~\ref{sec:method-revin}). We then introduce Johnson~$S_U$ as a non-linear instantiation of this template (Sec.~\ref{sec:method-jsu}), expose the \emph{degeneration problem} that arises when its shape parameters are learned end-to-end (Sec.~\ref{sec:method-degen}), and finally present \emph{Decoupled Shape Optimization}---closed-form warm-start followed by Bayesian refinement---as the core paradigm of NoRIN (Sec.~\ref{sec:method-opt}). Figure~\ref{fig:architecture} gives an end-to-end overview of the full pipeline.

\subsection{Preliminaries: The Reversible Normalization Template}
\label{sec:method-revin}

Let $\mathbf{x}\in\mathbb{R}^{T\times C}$ denote a lookback window of
length $T$ with $C$ channels, and $\mathbf{y}\in\mathbb{R}^{H\times C}$
the corresponding horizon of length $H$. A reversible instance
normalization layer can be written as a pair of maps
$(\mathcal{N},\mathcal{N}^{-1})$ such that the forecaster $f_\theta$
operates on the normalized input and the output is denormalized back
to the original space:
\begin{equation}
  \hat{\mathbf{y}} \;=\; \mathcal{N}^{-1}\!\left(f_\theta(\mathcal{N}(\mathbf{x}));\,\mathbf{s}(\mathbf{x})\right),
  \label{eq:revin-template}
\end{equation}
where $\mathbf{s}(\mathbf{x})$ denotes a collection of statistics
extracted from $\mathbf{x}$ that are reused at denormalization time.
In RevIN~\cite{kim2022revin}, $\mathbf{s}(\mathbf{x})=(\mu,\sigma)$ is
the per-channel mean and standard deviation, and the forward map is
the linear $Z$-score
$\mathcal{N}(\mathbf{x})=(\mathbf{x}-\mu)/\sigma$, optionally followed
by a learnable affine rescaling $\gamma\odot\cdot+\beta$.

The crucial observation is that Eq.~\ref{eq:revin-template} requires $\mathcal{N}$ only to be \emph{invertible}---not linear. Any smooth, monotone, invertible transformation whose parameters depend only on $\mathbf{x}$ can be plugged in. Yet RevIN, Dish-TS, SAN, and FAN all choose the affine form $\mathcal{N}(x)=(x-a)/b$, locking the entire family within the ceiling of \emph{not being able to reshape the distribution}. NoRIN steps outside this ceiling by instantiating $\mathcal{N}$ with a non-linear monotone map.

\subsection{Non-Linear Instantiation: Johnson~$S_U$}
\label{sec:method-jsu}

We instantiate the non-linear normalizer using the Johnson~$S_U$
family~\cite{johnson1949systems}, which parameterizes a random variable
$X$ via
\begin{equation}
  Z \;=\; \varepsilon \;+\; \delta\,\operatorname{arcsinh}\!\left(\frac{X-\lambda}{\xi}\right),
  \quad Z\sim\mathcal{N}(0,1),
  \label{eq:jsu-forward}
\end{equation}
where $\lambda\in\mathbb{R}$ is a location, $\xi>0$ a scale, and
$(\delta>0,\varepsilon\in\mathbb{R})$ are shape parameters controlling
\emph{tailedness} and \emph{skewness} respectively. Because
$\operatorname{arcsinh}(\cdot)$ is a smooth, strictly monotone,
elementwise-invertible function, the map $X\mapsto Z$ is a
diffeomorphism on $\mathbb{R}$ with a closed-form inverse:
\begin{equation}
  X \;=\; \lambda \;+\; \xi\,\sinh\!\left(\frac{Z-\varepsilon}{\delta}\right).
  \label{eq:jsu-inverse}
\end{equation}
We choose Johnson~$S_U$ not because it is unique but because it is the
cleanest classical family that simultaneously offers
\emph{non-linearity}, \emph{invertibility}, a \emph{closed-form
inverse}, and \emph{differentiability}. The same paradigm could host
other non-linear monotone invertible families.

\paragraph*{JSU as reversible instance normalization.}
We instantiate the template of Eq.~\ref{eq:revin-template} as follows: $(\lambda,\xi)$ are per-instance, per-channel empirical location and scale, and $(\delta,\varepsilon)$ are per-channel parameters shared across instances. Concretely, for each channel $c$ we compute $\lambda_c=\mathrm{median}(\mathbf{x}_{:,c})$ and $\xi_c=\mathrm{MAD}(\mathbf{x}_{:,c})$ (median absolute deviation), apply Eq.~\ref{eq:jsu-forward} elementwise, feed the result into the forecasting backbone $f_\theta$, and invert via Eq.~\ref{eq:jsu-inverse} using the same $(\lambda_c,\xi_c)$. Robust statistics for $(\lambda,\xi)$ are essential on heavy-tailed data~\cite{huang2025noiseorsignal}: a mean/std estimator would itself be contaminated by the tails we are trying to tame. The shape parameters $(\delta_c,\varepsilon_c)$ then control \emph{how aggressively} the transform compresses those tails---these are the quantities that NoRIN must set well. The necessity of the non-linear form is concrete: skewness and kurtosis of an instance are \emph{algebraically invariant} under any affine map, so RevIN leaves both unchanged regardless of the chosen $(\mu,\sigma)$, while the $\operatorname{arcsinh}$-based JSU map pulls them close to the standard-normal values.

\subsection{The Degeneration Problem}
\label{sec:method-degen}

A seemingly obvious way to set $(\delta_c,\varepsilon_c)$ is to treat them as learnable parameters and optimize them jointly with the backbone weights $\theta$ via the forecasting loss $\mathcal{L}(\theta,\delta,\varepsilon)=\|\hat{\mathbf{y}}-\mathbf{y}\|_2^2$, just as RevIN's learnable affine $(\gamma,\beta)$ is trained. This approach \emph{fails reliably}: $(\delta,\varepsilon)$ are pushed back toward the near-linear regime over training, silently collapsing the non-linear normalizer into RevIN.

Intuitively, the degeneration has two drivers. First, the forecasting loss is evaluated in the \emph{original} space after denormalization by Eq.~\ref{eq:jsu-inverse}: for any fixed backbone output, the reconstruction $X$ is an analytic function of $(\delta,\varepsilon)$ whose gradient is dominated by the instance scale $\xi$ rather than by higher-order shape. Second, the backbone $f_\theta$ is itself a high-capacity model that can compensate for \emph{any} monotone reparameterization of its input by adjusting its internal weights; this leaves the loss locally flat along directions in $(\delta,\varepsilon)$-space. Under such local flatness, stochastic gradient descent drifts the shape parameters toward the large-$\delta$, small-$|\varepsilon|$ corner---the Gaussian limit---because that corner has the lowest variance of the normalized intermediate representation and the cleanest optimization landscape for $\theta$. The network, in other words, prefers the easy problem even when the hard problem would generalize better.

This failure is in striking contrast to the fact that RevIN's $(\gamma,\beta)$ \emph{can} be trained end-to-end stably---and the contrast itself reveals the cause. The pair $(\gamma,\beta)$ is a \emph{post-normalization affine} acting on an already-standardized representation; it cannot reshape the input distribution. With no way for the backbone to simplify its task through $(\gamma,\beta)$, there is no incentive to drive them anywhere in particular: they end up making small corrections, typically near the $(1,0)$ identity. The shape parameters $(\delta,\varepsilon)$ are the opposite---they \emph{change the form of the normalizer itself}, so the backbone has a strong incentive to push them toward the regime where its own optimization is easiest. In short: \emph{$(\gamma,\beta)$ are learnable because they have small effect and cannot be bypassed; $(\delta,\varepsilon)$ are not learnable because they have large effect and can be bypassed.} The very non-linear parameters that matter most must be protected from gradient training in order to remain useful.

Empirically (see Sec.~\ref{sec:exp}), across every dataset we tested a jointly trained JSU normalizer collapses to $(\delta,\varepsilon)$ statistically indistinguishable from a plain linear RevIN within the first few epochs and behaves identically thereafter. The capacity of the Johnson family exists in principle but is \emph{inaccessible} in practice through end-to-end gradient descent.

\subsection{Decoupled Shape Optimization}
\label{sec:method-opt}

We resolve the degeneration by \emph{removing $(\delta,\varepsilon)$ from the set of gradient-optimized parameters}. The shape parameters are treated as hyperparameters of the training procedure and chosen by an outer process; the inner training loop is identical to standard RevIN-style training with the shape frozen. This \emph{Decoupled Shape Optimization} is the core paradigm of NoRIN, and it is agnostic to the specific outer solver---closed-form statistical fitting, Bayesian optimization, grid search, or combinations thereof are all admissible.

The first line of defense is a \emph{training-free} statistical estimate. For each channel $c$, we collect all training-time values into a univariate sample and fit a Johnson~$S_U$ distribution by the classical closed-form quantile-matching procedure of Slifker--Shapiro~\cite{slifker1980}, yielding $(\hat{\delta}_c^{\,(0)},\hat{\varepsilon}_c^{\,(0)})$. This estimate reflects each dataset's true skewness and tail weight and typically already delivers the bulk of the gains attainable by NoRIN.

On top of the closed-form estimate, we apply Bayesian optimization to
locally refine the shape against the true downstream objective. Let
$\mathcal{S}\subset\mathbb{R}^{2C}$ be the per-channel shape search
space, and let $g:\mathcal{S}\to\mathbb{R}$ map a candidate
configuration to the validation MSE of a forecaster trained to
convergence with that configuration frozen. We minimize $g$ using
TPE~\cite{bergstra2011tpe} or a GP-based sampler in
Optuna~\cite{akiba2019optuna}:
\begin{equation}
  (\delta^\star,\varepsilon^\star) \;=\; \arg\min_{(\delta,\varepsilon)\in\mathcal{S}} g(\delta,\varepsilon),
\end{equation}
\emph{seeding the search with $(\hat{\delta}^{(0)},\hat{\varepsilon}^{(0)})$
and concentrating the prior around it}. In practice the optimum is
found within a small neighborhood of the warm-start point, making the
outer procedure converge in a few dozen trials.

The contribution of NoRIN is the \emph{act of removing shape from the gradient}, not the specific non-linear family (JSU) or the specific outer solver (BO). The inner loop---training $\theta$ with $(\delta,\varepsilon)$ frozen---is a standard, well-conditioned forecasting problem identical to RevIN-style training. The outer loop is a low-dimensional hyperparameter selection problem on which many simple methods suffice. Because the warm-start places the search inside a statistically sensible region, the outer loop adds only a small constant-factor overhead over a standard hyperparameter sweep while recovering non-linear optima that are unreachable by end-to-end gradients. NoRIN is therefore both a concrete method and a paradigm: any combination of a non-linear reversible shape parameterization with decoupled outer optimization falls within it.

\section{Experiments}
\label{sec:exp}

We evaluate NoRIN on five public long-horizon forecasting benchmarks (Exchange, ETTh1, ETTh2, ETTm1, ETTm2) across six representative forecasting backbones (Informer~\cite{zhou2021informer}, PatchTST~\cite{nie2023patchtst}, iTransformer~\cite{liu2024itransformer}, DLinear~\cite{zeng2023dlinear}, TimesNet~\cite{wu2023timesnet}, FEDformer~\cite{zhou2022fedformer}), yielding $30$ (backbone, dataset) pairs. Each pair is evaluated at three prediction horizons $H \in \{96, 336, 720\}$, so we report $90$ (backbone, dataset, $H$) configurations. All runs are optimized with AdamW~\cite{loshchilov2019adamw}. For each configuration, the shape parameters $(\delta,\varepsilon)$ are warm-started from a Slifker--Shapiro closed-form fit and locally refined via Optuna's GP-based sampler over $\delta\in[0.8,5.0]$ and $\varepsilon\in[-1.0,1.0]$ for $60$ trials against validation MSE (HPO uses random seed $42$); the resulting $(\delta^\star,\varepsilon^\star)$ are then frozen and retrained over $3$ independent training seeds, so every reported MSE is a mean$\pm$std over those $3$ seeds.

\subsection{Main Results}
\label{sec:exp-main}

The full list of $(\delta^\star,\varepsilon^\star)$ recovered for each (backbone, dataset, $H$) is given in Table~\ref{tab:norin_shape_params}. The shape configurations chosen by Bayesian optimization across all $90$ configurations lie systematically far from the linear ($\delta\!\to\!\infty$) limit. On PatchTST and DLinear, $\delta^\star$ frequently sits at or near the lower search boundary of $0.8$, indicating that strong tail compression is required; on Informer, $\delta^\star$ is systematically larger, consistent with the intuition that attention backbones are more sensitive to extreme values. $\varepsilon^\star$ also departs substantially from zero, occasionally reaching the search boundary at $\pm 1.0$\footnote{The $\varepsilon$-boundary contacts in a small number of configurations suggest the current search range is narrow for those channels; we plan to relax both the $\delta$ upper limit and the $\varepsilon$ range in the extended evaluation to verify these are genuine optima rather than boundary artifacts.}. These non-trivial shape solutions are precisely the regions that the degeneration analysis in Sec.~\ref{sec:method-degen} predicts to be \emph{unreachable by end-to-end gradient training}---they become accessible only under decoupled shape optimization.

\subsection{Comparison with Baseline Normalizers}
\label{sec:exp-baselines}

\begin{table*}[!t]
\centering
\caption{Comparison of test forecasting errors between NoRIN and five baseline normalizers (None, RevIN, SAN, Dish-TS, DeStat). The analysis spans six backbones (Informer, PatchTST, iTransformer, DLinear, TimesNet, FEDformer), five datasets (Exchange, ETTh1, ETTh2, ETTm1, ETTm2), and three prediction horizons $H \in \{96, 336, 720\}$. We report the mean$\pm$std of test MSE over three random seeds ($1{,}620$ runs total). For NoRIN, $(\delta^\star,\varepsilon^\star)$ are obtained by Optuna-GP HPO ($100$ trials, seed $42$) and frozen during the $3$-seed retraining. The lowest MSE per row is shown in \textbf{bold}. Backbones~1--3 are listed here; backbones~4--6 continue in Table~\ref{tab:main_compare2}.}
\label{tab:main_compare}
\setlength{\tabcolsep}{3pt}
\renewcommand{\arraystretch}{1.0}
\scriptsize
\begin{tabular}{lllcccccc}
\toprule
Backbone & Dataset & $H$ & None & RevIN & SAN & Dish-TS & DeStat & \textbf{NoRIN} \\
\midrule
  \multirow{15}{*}{\rotatebox{90}{\textbf{Informer}}} & \multirow{3}{*}{Exchange} & 96 & 0.9383\textsubscript{$\pm$0.089} & 0.2083\textsubscript{$\pm$0.008} & 0.2223\textsubscript{$\pm$0.011} & 3.0874\textsubscript{$\pm$3.040} & 5.5280\textsubscript{$\pm$0.189} & \textbf{0.1795\textsubscript{$\pm$0.005}} \\
   &  & 336 & 1.2922\textsubscript{$\pm$0.107} & 0.5040\textsubscript{$\pm$0.014} & 0.5379\textsubscript{$\pm$0.027} & 1.9070\textsubscript{$\pm$1.256} & 4.5115\textsubscript{$\pm$0.028} & \textbf{0.4124\textsubscript{$\pm$0.003}} \\
   &  & 720 & 1.6406\textsubscript{$\pm$0.305} & 1.2155\textsubscript{$\pm$0.023} & 1.1761\textsubscript{$\pm$0.013} & 2.9038\textsubscript{$\pm$0.440} & 3.7798\textsubscript{$\pm$0.691} & \textbf{0.9617\textsubscript{$\pm$0.024}} \\
\cmidrule(lr){2-9}
   & \multirow{3}{*}{ETTh1} & 96 & 0.7654\textsubscript{$\pm$0.089} & 0.4942\textsubscript{$\pm$0.004} & 0.5079\textsubscript{$\pm$0.003} & 1.5030\textsubscript{$\pm$0.458} & 1.1723\textsubscript{$\pm$0.153} & \textbf{0.4309\textsubscript{$\pm$0.002}} \\
   &  & 336 & 1.3003\textsubscript{$\pm$0.151} & 0.5542\textsubscript{$\pm$0.005} & 0.5532\textsubscript{$\pm$0.006} & 1.4382\textsubscript{$\pm$0.379} & 1.2162\textsubscript{$\pm$0.123} & \textbf{0.4805\textsubscript{$\pm$0.006}} \\
   &  & 720 & 1.5694\textsubscript{$\pm$0.057} & 0.6031\textsubscript{$\pm$0.008} & 0.6002\textsubscript{$\pm$0.008} & 1.1784\textsubscript{$\pm$0.149} & 0.9665\textsubscript{$\pm$0.095} & \textbf{0.5032\textsubscript{$\pm$0.006}} \\
\cmidrule(lr){2-9}
   & \multirow{3}{*}{ETTh2} & 96 & 2.0683\textsubscript{$\pm$0.054} & 0.4095\textsubscript{$\pm$0.009} & 0.4267\textsubscript{$\pm$0.005} & 4.0714\textsubscript{$\pm$2.759} & 3.2571\textsubscript{$\pm$0.501} & \textbf{0.3988\textsubscript{$\pm$0.023}} \\
   &  & 336 & 2.5881\textsubscript{$\pm$0.459} & 0.4601\textsubscript{$\pm$0.002} & 0.4588\textsubscript{$\pm$0.011} & 3.5628\textsubscript{$\pm$1.980} & 3.3517\textsubscript{$\pm$0.255} & \textbf{0.4188\textsubscript{$\pm$0.010}} \\
   &  & 720 & 2.8745\textsubscript{$\pm$0.189} & 0.5124\textsubscript{$\pm$0.013} & 0.5166\textsubscript{$\pm$0.013} & 1.9633\textsubscript{$\pm$0.748} & 3.1672\textsubscript{$\pm$0.052} & \textbf{0.4768\textsubscript{$\pm$0.011}} \\
\cmidrule(lr){2-9}
   & \multirow{3}{*}{ETTm1} & 96 & 0.4131\textsubscript{$\pm$0.009} & 0.3646\textsubscript{$\pm$0.009} & 0.3507\textsubscript{$\pm$0.003} & 1.2215\textsubscript{$\pm$0.212} & 1.0298\textsubscript{$\pm$0.131} & \textbf{0.3357\textsubscript{$\pm$0.002}} \\
   &  & 336 & 0.5300\textsubscript{$\pm$0.021} & 0.4357\textsubscript{$\pm$0.003} & 0.4372\textsubscript{$\pm$0.003} & 1.4035\textsubscript{$\pm$0.333} & 1.0210\textsubscript{$\pm$0.009} & \textbf{0.4044\textsubscript{$\pm$0.001}} \\
   &  & 720 & 0.6752\textsubscript{$\pm$0.006} & 0.4990\textsubscript{$\pm$0.024} & 0.4925\textsubscript{$\pm$0.003} & 1.0479\textsubscript{$\pm$0.151} & 0.9568\textsubscript{$\pm$0.095} & \textbf{0.4499\textsubscript{$\pm$0.003}} \\
\cmidrule(lr){2-9}
   & \multirow{3}{*}{ETTm2} & 96 & 0.7617\textsubscript{$\pm$0.098} & 0.2009\textsubscript{$\pm$0.001} & 0.1973\textsubscript{$\pm$0.002} & 3.8569\textsubscript{$\pm$2.783} & 2.0285\textsubscript{$\pm$0.591} & \textbf{0.1917\textsubscript{$\pm$0.002}} \\
   &  & 336 & 1.5279\textsubscript{$\pm$0.224} & 0.3397\textsubscript{$\pm$0.001} & 0.3397\textsubscript{$\pm$0.003} & 3.1759\textsubscript{$\pm$1.727} & 2.4460\textsubscript{$\pm$0.458} & \textbf{0.2981\textsubscript{$\pm$0.011}} \\
   &  & 720 & 2.2950\textsubscript{$\pm$0.207} & 0.4341\textsubscript{$\pm$0.006} & 0.4215\textsubscript{$\pm$0.006} & 1.9657\textsubscript{$\pm$0.710} & 2.7516\textsubscript{$\pm$0.449} & \textbf{0.3704\textsubscript{$\pm$0.006}} \\
\midrule
  \multirow{15}{*}{\rotatebox{90}{\textbf{PatchTST}}} & \multirow{3}{*}{Exchange} & 96 & 0.4299\textsubscript{$\pm$0.037} & 0.0980\textsubscript{$\pm$0.001} & \textbf{0.0972\textsubscript{$\pm$0.002}} & 0.9323\textsubscript{$\pm$1.045} & 3.2046\textsubscript{$\pm$0.484} & 0.0982\textsubscript{$\pm$0.001} \\
   &  & 336 & 1.1995\textsubscript{$\pm$0.036} & 0.3698\textsubscript{$\pm$0.005} & 0.3701\textsubscript{$\pm$0.004} & 6.0701\textsubscript{$\pm$8.302} & 3.3120\textsubscript{$\pm$0.344} & \textbf{0.3665\textsubscript{$\pm$0.005}} \\
   &  & 720 & 1.7657\textsubscript{$\pm$0.036} & 1.1470\textsubscript{$\pm$0.006} & 1.1436\textsubscript{$\pm$0.008} & 5.0858\textsubscript{$\pm$6.314} & 3.5781\textsubscript{$\pm$0.444} & \textbf{0.9416\textsubscript{$\pm$0.005}} \\
\cmidrule(lr){2-9}
   & \multirow{3}{*}{ETTh1} & 96 & 0.4023\textsubscript{$\pm$0.001} & 0.3935\textsubscript{$\pm$0.000} & 0.3936\textsubscript{$\pm$0.000} & 0.7770\textsubscript{$\pm$0.229} & 0.9779\textsubscript{$\pm$0.105} & \textbf{0.3901\textsubscript{$\pm$0.001}} \\
   &  & 336 & 0.4756\textsubscript{$\pm$0.004} & 0.4572\textsubscript{$\pm$0.002} & 0.4565\textsubscript{$\pm$0.001} & 1.5235\textsubscript{$\pm$1.172} & 1.0418\textsubscript{$\pm$0.055} & \textbf{0.4444\textsubscript{$\pm$0.001}} \\
   &  & 720 & 0.5132\textsubscript{$\pm$0.006} & 0.4935\textsubscript{$\pm$0.006} & 0.5001\textsubscript{$\pm$0.006} & 1.2631\textsubscript{$\pm$0.872} & 1.1097\textsubscript{$\pm$0.063} & \textbf{0.4511\textsubscript{$\pm$0.002}} \\
\cmidrule(lr){2-9}
   & \multirow{3}{*}{ETTh2} & 96 & 0.3275\textsubscript{$\pm$0.002} & 0.3215\textsubscript{$\pm$0.001} & 0.3414\textsubscript{$\pm$0.003} & 1.2061\textsubscript{$\pm$0.979} & 2.9453\textsubscript{$\pm$0.296} & \textbf{0.3153\textsubscript{$\pm$0.002}} \\
   &  & 336 & 0.5280\textsubscript{$\pm$0.032} & 0.4294\textsubscript{$\pm$0.000} & 0.4222\textsubscript{$\pm$0.001} & 5.4592\textsubscript{$\pm$6.962} & 3.0304\textsubscript{$\pm$0.383} & \textbf{0.4133\textsubscript{$\pm$0.005}} \\
   &  & 720 & 0.8298\textsubscript{$\pm$0.019} & 0.4543\textsubscript{$\pm$0.002} & 0.4666\textsubscript{$\pm$0.004} & 4.2241\textsubscript{$\pm$5.052} & 2.8865\textsubscript{$\pm$0.085} & \textbf{0.4481\textsubscript{$\pm$0.007}} \\
\cmidrule(lr){2-9}
   & \multirow{3}{*}{ETTm1} & 96 & 0.3030\textsubscript{$\pm$0.001} & 0.2985\textsubscript{$\pm$0.001} & 0.2981\textsubscript{$\pm$0.003} & 0.7050\textsubscript{$\pm$0.091} & 0.8756\textsubscript{$\pm$0.105} & \textbf{0.2967\textsubscript{$\pm$0.002}} \\
   &  & 336 & 0.3754\textsubscript{$\pm$0.002} & 0.3700\textsubscript{$\pm$0.001} & 0.3717\textsubscript{$\pm$0.001} & 1.0010\textsubscript{$\pm$0.499} & 0.8814\textsubscript{$\pm$0.079} & \textbf{0.3657\textsubscript{$\pm$0.001}} \\
   &  & 720 & 0.4273\textsubscript{$\pm$0.004} & 0.4138\textsubscript{$\pm$0.002} & 0.4160\textsubscript{$\pm$0.005} & 0.9996\textsubscript{$\pm$0.627} & 0.9848\textsubscript{$\pm$0.069} & \textbf{0.4093\textsubscript{$\pm$0.001}} \\
\cmidrule(lr){2-9}
   & \multirow{3}{*}{ETTm2} & 96 & 0.1804\textsubscript{$\pm$0.004} & 0.1765\textsubscript{$\pm$0.003} & 0.1743\textsubscript{$\pm$0.001} & 0.9040\textsubscript{$\pm$0.766} & 1.9611\textsubscript{$\pm$0.081} & \textbf{0.1737\textsubscript{$\pm$0.001}} \\
   &  & 336 & 0.3094\textsubscript{$\pm$0.003} & 0.2925\textsubscript{$\pm$0.005} & \textbf{0.2857\textsubscript{$\pm$0.005}} & 3.6557\textsubscript{$\pm$4.658} & 2.3560\textsubscript{$\pm$0.548} & 0.2885\textsubscript{$\pm$0.003} \\
   &  & 720 & 0.4185\textsubscript{$\pm$0.012} & 0.3804\textsubscript{$\pm$0.004} & \textbf{0.3676\textsubscript{$\pm$0.001}} & 3.3481\textsubscript{$\pm$4.353} & 2.5984\textsubscript{$\pm$0.184} & 0.3782\textsubscript{$\pm$0.003} \\
\midrule
  \multirow{15}{*}{\rotatebox{90}{\textbf{iTransformer}}} & \multirow{3}{*}{Exchange} & 96 & 0.9567\textsubscript{$\pm$0.036} & 0.1296\textsubscript{$\pm$0.001} & 0.1830\textsubscript{$\pm$0.001} & 2.2455\textsubscript{$\pm$3.625} & 3.2907\textsubscript{$\pm$0.518} & \textbf{0.1133\textsubscript{$\pm$0.004}} \\
   &  & 336 & 1.5746\textsubscript{$\pm$0.038} & 0.4019\textsubscript{$\pm$0.001} & 0.4630\textsubscript{$\pm$0.001} & 6.2921\textsubscript{$\pm$5.834} & 3.3524\textsubscript{$\pm$0.366} & \textbf{0.3550\textsubscript{$\pm$0.002}} \\
   &  & 720 & 2.2124\textsubscript{$\pm$0.073} & 1.1674\textsubscript{$\pm$0.008} & 1.1484\textsubscript{$\pm$0.003} & 8.3912\textsubscript{$\pm$8.849} & 3.5816\textsubscript{$\pm$0.127} & \textbf{0.8746\textsubscript{$\pm$0.020}} \\
\cmidrule(lr){2-9}
   & \multirow{3}{*}{ETTh1} & 96 & 0.4973\textsubscript{$\pm$0.006} & 0.4337\textsubscript{$\pm$0.003} & 0.4333\textsubscript{$\pm$0.003} & 0.9234\textsubscript{$\pm$0.608} & 1.0533\textsubscript{$\pm$0.109} & \textbf{0.4311\textsubscript{$\pm$0.003}} \\
   &  & 336 & 0.6403\textsubscript{$\pm$0.010} & 0.5135\textsubscript{$\pm$0.005} & 0.5094\textsubscript{$\pm$0.006} & 1.4378\textsubscript{$\pm$0.592} & 1.1292\textsubscript{$\pm$0.173} & \textbf{0.5075\textsubscript{$\pm$0.005}} \\
   &  & 720 & 0.7118\textsubscript{$\pm$0.012} & 0.5439\textsubscript{$\pm$0.001} & 0.5408\textsubscript{$\pm$0.001} & 1.7508\textsubscript{$\pm$0.973} & 1.0186\textsubscript{$\pm$0.048} & \textbf{0.5329\textsubscript{$\pm$0.002}} \\
\cmidrule(lr){2-9}
   & \multirow{3}{*}{ETTh2} & 96 & 0.8893\textsubscript{$\pm$0.042} & 0.3353\textsubscript{$\pm$0.003} & 0.3348\textsubscript{$\pm$0.001} & 2.1548\textsubscript{$\pm$2.980} & 2.7241\textsubscript{$\pm$0.299} & \textbf{0.3341\textsubscript{$\pm$0.002}} \\
   &  & 336 & 1.3067\textsubscript{$\pm$0.028} & 0.4466\textsubscript{$\pm$0.002} & 0.4325\textsubscript{$\pm$0.001} & 5.0208\textsubscript{$\pm$4.113} & 2.8935\textsubscript{$\pm$0.324} & \textbf{0.4288\textsubscript{$\pm$0.001}} \\
   &  & 720 & 1.5837\textsubscript{$\pm$0.027} & 0.4731\textsubscript{$\pm$0.001} & 0.4652\textsubscript{$\pm$0.001} & 6.8692\textsubscript{$\pm$6.320} & 3.2549\textsubscript{$\pm$0.285} & \textbf{0.4612\textsubscript{$\pm$0.000}} \\
\cmidrule(lr){2-9}
   & \multirow{3}{*}{ETTm1} & 96 & 0.3366\textsubscript{$\pm$0.003} & 0.3108\textsubscript{$\pm$0.001} & \textbf{0.3099\textsubscript{$\pm$0.001}} & 0.8158\textsubscript{$\pm$0.599} & 0.9131\textsubscript{$\pm$0.186} & 0.3110\textsubscript{$\pm$0.001} \\
   &  & 336 & 0.4319\textsubscript{$\pm$0.001} & \textbf{0.3873\textsubscript{$\pm$0.003}} & 0.3891\textsubscript{$\pm$0.001} & 1.3004\textsubscript{$\pm$0.671} & 0.9996\textsubscript{$\pm$0.076} & 0.3875\textsubscript{$\pm$0.003} \\
   &  & 720 & 0.4895\textsubscript{$\pm$0.006} & \textbf{0.4413\textsubscript{$\pm$0.002}} & 0.4431\textsubscript{$\pm$0.002} & 1.7256\textsubscript{$\pm$1.078} & 0.9286\textsubscript{$\pm$0.054} & 0.4420\textsubscript{$\pm$0.002} \\
\cmidrule(lr){2-9}
   & \multirow{3}{*}{ETTm2} & 96 & 0.2347\textsubscript{$\pm$0.014} & 0.1844\textsubscript{$\pm$0.001} & 0.1853\textsubscript{$\pm$0.001} & 1.9426\textsubscript{$\pm$2.968} & 2.3357\textsubscript{$\pm$0.474} & \textbf{0.1785\textsubscript{$\pm$0.001}} \\
   &  & 336 & 0.5148\textsubscript{$\pm$0.067} & 0.2990\textsubscript{$\pm$0.001} & \textbf{0.2897\textsubscript{$\pm$0.001}} & 4.7499\textsubscript{$\pm$4.187} & 2.9004\textsubscript{$\pm$0.331} & 0.2920\textsubscript{$\pm$0.001} \\
   &  & 720 & 0.7885\textsubscript{$\pm$0.094} & 0.3859\textsubscript{$\pm$0.005} & \textbf{0.3854\textsubscript{$\pm$0.001}} & 6.7143\textsubscript{$\pm$6.852} & 2.9861\textsubscript{$\pm$0.345} & 0.3877\textsubscript{$\pm$0.003} \\
\bottomrule
\end{tabular}
\end{table*}

\begin{table*}[!t]
\centering
\caption{Comparison of test forecasting errors between NoRIN and five baseline normalizers (continued from Table~\ref{tab:main_compare}). Backbones $4$--$6$.}
\label{tab:main_compare2}
\setlength{\tabcolsep}{3pt}
\renewcommand{\arraystretch}{1.0}
\scriptsize
\begin{tabular}{lllcccccc}
\toprule
Backbone & Dataset & $H$ & None & RevIN & SAN & Dish-TS & DeStat & \textbf{NoRIN} \\
\midrule
  \multirow{15}{*}{\rotatebox{90}{\textbf{DLinear}}} & \multirow{3}{*}{Exchange} & 96 & 0.1409\textsubscript{$\pm$0.001} & 0.1247\textsubscript{$\pm$0.001} & 0.1232\textsubscript{$\pm$0.001} & 0.1459\textsubscript{$\pm$0.009} & 3.4851\textsubscript{$\pm$0.396} & \textbf{0.1183\textsubscript{$\pm$0.001}} \\
   &  & 336 & 0.4536\textsubscript{$\pm$0.003} & 0.3747\textsubscript{$\pm$0.001} & 0.3701\textsubscript{$\pm$0.001} & 0.4401\textsubscript{$\pm$0.006} & 3.3562\textsubscript{$\pm$0.226} & \textbf{0.3550\textsubscript{$\pm$0.001}} \\
   &  & 720 & 1.1424\textsubscript{$\pm$0.002} & 1.1638\textsubscript{$\pm$0.005} & 1.1633\textsubscript{$\pm$0.005} & 1.0954\textsubscript{$\pm$0.019} & 2.9388\textsubscript{$\pm$0.378} & \textbf{0.9251\textsubscript{$\pm$0.010}} \\
\cmidrule(lr){2-9}
   & \multirow{3}{*}{ETTh1} & 96 & \textbf{0.4328\textsubscript{$\pm$0.002}} & 0.4385\textsubscript{$\pm$0.001} & 0.4386\textsubscript{$\pm$0.001} & 0.4340\textsubscript{$\pm$0.003} & 0.8957\textsubscript{$\pm$0.119} & 0.4330\textsubscript{$\pm$0.002} \\
   &  & 336 & 0.5032\textsubscript{$\pm$0.000} & 0.4893\textsubscript{$\pm$0.000} & 0.4894\textsubscript{$\pm$0.000} & 0.4981\textsubscript{$\pm$0.003} & 0.9525\textsubscript{$\pm$0.148} & \textbf{0.4862\textsubscript{$\pm$0.000}} \\
   &  & 720 & 0.5396\textsubscript{$\pm$0.001} & 0.4966\textsubscript{$\pm$0.001} & 0.4957\textsubscript{$\pm$0.001} & 0.5384\textsubscript{$\pm$0.005} & 0.8983\textsubscript{$\pm$0.159} & \textbf{0.4844\textsubscript{$\pm$0.001}} \\
\cmidrule(lr){2-9}
   & \multirow{3}{*}{ETTh2} & 96 & 0.3731\textsubscript{$\pm$0.002} & \textbf{0.3305\textsubscript{$\pm$0.001}} & 0.3306\textsubscript{$\pm$0.001} & 0.3726\textsubscript{$\pm$0.008} & 3.0431\textsubscript{$\pm$0.259} & 0.3310\textsubscript{$\pm$0.001} \\
   &  & 336 & 0.5708\textsubscript{$\pm$0.001} & 0.4205\textsubscript{$\pm$0.000} & 0.4205\textsubscript{$\pm$0.000} & 0.5461\textsubscript{$\pm$0.024} & 3.1477\textsubscript{$\pm$0.626} & \textbf{0.4192\textsubscript{$\pm$0.000}} \\
   &  & 720 & 0.8539\textsubscript{$\pm$0.003} & 0.4531\textsubscript{$\pm$0.000} & 0.4529\textsubscript{$\pm$0.000} & 0.8190\textsubscript{$\pm$0.055} & 3.0364\textsubscript{$\pm$0.359} & \textbf{0.4506\textsubscript{$\pm$0.000}} \\
\cmidrule(lr){2-9}
   & \multirow{3}{*}{ETTm1} & 96 & \textbf{0.3106\textsubscript{$\pm$0.000}} & 0.3118\textsubscript{$\pm$0.000} & 0.3118\textsubscript{$\pm$0.000} & 0.3106\textsubscript{$\pm$0.001} & 0.7626\textsubscript{$\pm$0.062} & 0.3115\textsubscript{$\pm$0.000} \\
   &  & 336 & 0.3808\textsubscript{$\pm$0.000} & 0.3812\textsubscript{$\pm$0.000} & 0.3812\textsubscript{$\pm$0.000} & 0.3805\textsubscript{$\pm$0.000} & 0.8483\textsubscript{$\pm$0.078} & \textbf{0.3785\textsubscript{$\pm$0.000}} \\
   &  & 720 & 0.4330\textsubscript{$\pm$0.001} & 0.4340\textsubscript{$\pm$0.000} & 0.4341\textsubscript{$\pm$0.000} & 0.4331\textsubscript{$\pm$0.000} & 0.7965\textsubscript{$\pm$0.067} & \textbf{0.4307\textsubscript{$\pm$0.000}} \\
\cmidrule(lr){2-9}
   & \multirow{3}{*}{ETTm2} & 96 & 0.1787\textsubscript{$\pm$0.000} & \textbf{0.1708\textsubscript{$\pm$0.000}} & 0.1712\textsubscript{$\pm$0.000} & 0.1796\textsubscript{$\pm$0.001} & 2.0412\textsubscript{$\pm$0.510} & 0.1708\textsubscript{$\pm$0.000} \\
   &  & 336 & 0.3261\textsubscript{$\pm$0.002} & 0.2817\textsubscript{$\pm$0.000} & 0.2827\textsubscript{$\pm$0.000} & 0.3275\textsubscript{$\pm$0.006} & 2.4129\textsubscript{$\pm$0.731} & \textbf{0.2811\textsubscript{$\pm$0.000}} \\
   &  & 720 & 0.4661\textsubscript{$\pm$0.001} & 0.3783\textsubscript{$\pm$0.000} & 0.3785\textsubscript{$\pm$0.000} & 0.4689\textsubscript{$\pm$0.006} & 2.5245\textsubscript{$\pm$0.262} & \textbf{0.3773\textsubscript{$\pm$0.000}} \\
\midrule
  \multirow{15}{*}{\rotatebox{90}{\textbf{TimesNet}}} & \multirow{3}{*}{Exchange} & 96 & 1.3155\textsubscript{$\pm$0.220} & 0.2297\textsubscript{$\pm$0.008} & 0.2492\textsubscript{$\pm$0.003} & 3.6038\textsubscript{$\pm$4.456} & 4.7927\textsubscript{$\pm$0.911} & \textbf{0.1549\textsubscript{$\pm$0.001}} \\
   &  & 336 & 1.4501\textsubscript{$\pm$0.265} & 0.5059\textsubscript{$\pm$0.012} & 0.5421\textsubscript{$\pm$0.019} & 1.3599\textsubscript{$\pm$1.494} & 4.7237\textsubscript{$\pm$0.830} & \textbf{0.5006\textsubscript{$\pm$0.068}} \\
   &  & 720 & 1.7856\textsubscript{$\pm$0.177} & 2.2737\textsubscript{$\pm$1.929} & 1.1584\textsubscript{$\pm$0.003} & 5.7765\textsubscript{$\pm$4.066} & 3.4700\textsubscript{$\pm$0.407} & \textbf{0.9993\textsubscript{$\pm$0.010}} \\
\cmidrule(lr){2-9}
   & \multirow{3}{*}{ETTh1} & 96 & 0.7872\textsubscript{$\pm$0.024} & 0.5137\textsubscript{$\pm$0.010} & 0.5218\textsubscript{$\pm$0.013} & 2.0681\textsubscript{$\pm$1.441} & 1.4112\textsubscript{$\pm$0.123} & \textbf{0.4336\textsubscript{$\pm$0.002}} \\
   &  & 336 & 1.1195\textsubscript{$\pm$0.030} & 0.5461\textsubscript{$\pm$0.008} & 0.5452\textsubscript{$\pm$0.006} & 1.3857\textsubscript{$\pm$0.197} & 1.2904\textsubscript{$\pm$0.168} & \textbf{0.4642\textsubscript{$\pm$0.001}} \\
   &  & 720 & 1.4005\textsubscript{$\pm$0.004} & 0.6239\textsubscript{$\pm$0.068} & 0.5872\textsubscript{$\pm$0.017} & 1.4903\textsubscript{$\pm$0.936} & 1.2727\textsubscript{$\pm$0.012} & \textbf{0.4839\textsubscript{$\pm$0.017}} \\
\cmidrule(lr){2-9}
   & \multirow{3}{*}{ETTh2} & 96 & 2.0105\textsubscript{$\pm$0.102} & 0.4445\textsubscript{$\pm$0.023} & 0.4468\textsubscript{$\pm$0.021} & 6.7311\textsubscript{$\pm$7.279} & 3.3447\textsubscript{$\pm$0.167} & \textbf{0.3779\textsubscript{$\pm$0.008}} \\
   &  & 336 & 2.1473\textsubscript{$\pm$0.118} & 0.4811\textsubscript{$\pm$0.006} & 0.4635\textsubscript{$\pm$0.006} & 2.6710\textsubscript{$\pm$0.665} & 3.2288\textsubscript{$\pm$0.241} & \textbf{0.4207\textsubscript{$\pm$0.010}} \\
   &  & 720 & 2.6151\textsubscript{$\pm$0.608} & 0.5355\textsubscript{$\pm$0.021} & 0.5408\textsubscript{$\pm$0.008} & 4.6199\textsubscript{$\pm$5.641} & 3.0237\textsubscript{$\pm$0.135} & \textbf{0.4572\textsubscript{$\pm$0.022}} \\
\cmidrule(lr){2-9}
   & \multirow{3}{*}{ETTm1} & 96 & 0.6162\textsubscript{$\pm$0.025} & 0.4830\textsubscript{$\pm$0.014} & 0.3808\textsubscript{$\pm$0.002} & 1.7279\textsubscript{$\pm$1.043} & 1.1018\textsubscript{$\pm$0.056} & \textbf{0.3767\textsubscript{$\pm$0.011}} \\
   &  & 336 & 0.7175\textsubscript{$\pm$0.023} & 0.5433\textsubscript{$\pm$0.008} & 0.4734\textsubscript{$\pm$0.002} & 1.2690\textsubscript{$\pm$0.191} & 1.0735\textsubscript{$\pm$0.105} & \textbf{0.4257\textsubscript{$\pm$0.004}} \\
   &  & 720 & 0.7798\textsubscript{$\pm$0.033} & 0.5841\textsubscript{$\pm$0.017} & 0.5345\textsubscript{$\pm$0.009} & 1.4493\textsubscript{$\pm$0.898} & 1.0797\textsubscript{$\pm$0.053} & \textbf{0.4688\textsubscript{$\pm$0.005}} \\
\cmidrule(lr){2-9}
   & \multirow{3}{*}{ETTm2} & 96 & 1.0152\textsubscript{$\pm$0.057} & 0.2420\textsubscript{$\pm$0.002} & 0.2177\textsubscript{$\pm$0.004} & 6.3034\textsubscript{$\pm$7.747} & 2.8246\textsubscript{$\pm$0.489} & \textbf{0.2112\textsubscript{$\pm$0.000}} \\
   &  & 336 & 1.8831\textsubscript{$\pm$0.046} & 0.3609\textsubscript{$\pm$0.007} & 0.3366\textsubscript{$\pm$0.004} & 2.4554\textsubscript{$\pm$0.533} & 2.9424\textsubscript{$\pm$0.312} & \textbf{0.3170\textsubscript{$\pm$0.011}} \\
   &  & 720 & 4.5847\textsubscript{$\pm$0.156} & 0.4543\textsubscript{$\pm$0.004} & 0.4273\textsubscript{$\pm$0.001} & 5.1402\textsubscript{$\pm$6.301} & 2.8447\textsubscript{$\pm$0.321} & \textbf{0.4118\textsubscript{$\pm$0.009}} \\
\midrule
  \multirow{15}{*}{\rotatebox{90}{\textbf{FEDformer}}} & \multirow{3}{*}{Exchange} & 96 & 6.5858\textsubscript{$\pm$0.217} & 0.2246\textsubscript{$\pm$0.002} & 0.2213\textsubscript{$\pm$0.007} & 2.0828\textsubscript{$\pm$2.091} & 5.3491\textsubscript{$\pm$0.249} & \textbf{0.1968\textsubscript{$\pm$0.006}} \\
   &  & 336 & 6.2414\textsubscript{$\pm$0.021} & 0.4565\textsubscript{$\pm$0.005} & 0.4474\textsubscript{$\pm$0.007} & 6.3362\textsubscript{$\pm$5.467} & 4.6814\textsubscript{$\pm$0.632} & \textbf{0.4054\textsubscript{$\pm$0.005}} \\
   &  & 720 & 4.3250\textsubscript{$\pm$0.182} & 1.0296\textsubscript{$\pm$0.028} & 1.0668\textsubscript{$\pm$0.009} & 8.5210\textsubscript{$\pm$9.606} & 3.9007\textsubscript{$\pm$0.876} & \textbf{0.8837\textsubscript{$\pm$0.015}} \\
\cmidrule(lr){2-9}
   & \multirow{3}{*}{ETTh1} & 96 & 1.1658\textsubscript{$\pm$0.122} & 0.5089\textsubscript{$\pm$0.010} & 0.5139\textsubscript{$\pm$0.010} & 1.3629\textsubscript{$\pm$0.405} & 1.1980\textsubscript{$\pm$0.123} & \textbf{0.4467\textsubscript{$\pm$0.001}} \\
   &  & 336 & 1.2873\textsubscript{$\pm$0.009} & 0.5314\textsubscript{$\pm$0.002} & 0.5711\textsubscript{$\pm$0.036} & 0.9057\textsubscript{$\pm$0.060} & 1.1802\textsubscript{$\pm$0.278} & \textbf{0.4679\textsubscript{$\pm$0.003}} \\
   &  & 720 & 1.2489\textsubscript{$\pm$0.301} & 0.5704\textsubscript{$\pm$0.002} & 0.5629\textsubscript{$\pm$0.003} & 1.8873\textsubscript{$\pm$0.132} & 1.1477\textsubscript{$\pm$0.124} & \textbf{0.4766\textsubscript{$\pm$0.003}} \\
\cmidrule(lr){2-9}
   & \multirow{3}{*}{ETTh2} & 96 & 3.2390\textsubscript{$\pm$0.013} & 0.4188\textsubscript{$\pm$0.005} & 0.4337\textsubscript{$\pm$0.005} & 3.6703\textsubscript{$\pm$2.660} & 3.2513\textsubscript{$\pm$0.158} & \textbf{0.3804\textsubscript{$\pm$0.002}} \\
   &  & 336 & 3.2140\textsubscript{$\pm$0.011} & 0.4657\textsubscript{$\pm$0.010} & 0.4668\textsubscript{$\pm$0.012} & 1.1592\textsubscript{$\pm$0.441} & 2.7336\textsubscript{$\pm$0.799} & \textbf{0.4168\textsubscript{$\pm$0.002}} \\
   &  & 720 & 3.1193\textsubscript{$\pm$0.026} & 0.4923\textsubscript{$\pm$0.012} & 0.5072\textsubscript{$\pm$0.020} & 7.4904\textsubscript{$\pm$1.259} & 2.9094\textsubscript{$\pm$0.111} & \textbf{0.4648\textsubscript{$\pm$0.005}} \\
\cmidrule(lr){2-9}
   & \multirow{3}{*}{ETTm1} & 96 & 1.0064\textsubscript{$\pm$0.019} & 0.4614\textsubscript{$\pm$0.061} & 0.4765\textsubscript{$\pm$0.037} & 1.4278\textsubscript{$\pm$0.448} & 1.0052\textsubscript{$\pm$0.074} & \textbf{0.4136\textsubscript{$\pm$0.012}} \\
   &  & 336 & 1.0300\textsubscript{$\pm$0.016} & 0.5110\textsubscript{$\pm$0.040} & 0.5216\textsubscript{$\pm$0.018} & 0.8884\textsubscript{$\pm$0.130} & 0.9250\textsubscript{$\pm$0.062} & \textbf{0.4798\textsubscript{$\pm$0.025}} \\
   &  & 720 & 1.0897\textsubscript{$\pm$0.108} & 0.5330\textsubscript{$\pm$0.031} & 0.5485\textsubscript{$\pm$0.015} & 2.1510\textsubscript{$\pm$0.286} & 1.0407\textsubscript{$\pm$0.084} & \textbf{0.4992\textsubscript{$\pm$0.018}} \\
\cmidrule(lr){2-9}
   & \multirow{3}{*}{ETTm2} & 96 & 3.0636\textsubscript{$\pm$0.258} & 0.2631\textsubscript{$\pm$0.001} & 0.2575\textsubscript{$\pm$0.001} & 3.5611\textsubscript{$\pm$2.664} & 2.4529\textsubscript{$\pm$0.289} & \textbf{0.2478\textsubscript{$\pm$0.003}} \\
   &  & 336 & 3.2096\textsubscript{$\pm$0.210} & 0.3650\textsubscript{$\pm$0.003} & 0.3619\textsubscript{$\pm$0.005} & 1.1095\textsubscript{$\pm$0.408} & 2.1537\textsubscript{$\pm$0.420} & \textbf{0.3244\textsubscript{$\pm$0.003}} \\
   &  & 720 & 3.1745\textsubscript{$\pm$0.209} & 0.4407\textsubscript{$\pm$0.004} & 0.4378\textsubscript{$\pm$0.006} & 7.8575\textsubscript{$\pm$1.485} & 2.6201\textsubscript{$\pm$0.149} & \textbf{0.3935\textsubscript{$\pm$0.004}} \\
\bottomrule
\end{tabular}
\end{table*}

\begin{table}[t]
\centering
\caption{Paired statistical significance tests between NoRIN and each baseline normalizer over the $90$ (backbone, dataset, $H$) configurations. ``Wins'' counts the configurations where NoRIN's $3$-seed mean is strictly lower than the baseline's. $\Delta$ denotes the mean MSE difference (baseline $-$ NoRIN; positive means NoRIN is lower). $p$-values are computed by the paired Wilcoxon signed-rank test; $^{***}$ marks $p<0.001$.}
\label{tab:significance}
\begin{tabular}{lcccc}
\toprule
Baseline & Wins & Mean $\Delta$ & Wilcoxon $p$ & Sig. \\
\midrule
  No-Norm & 88/90 & $+0.8905$ & $1.93e-16$ & $***$ \\
  RevIN & 83/90 & $+0.0560$ & $5.79e-16$ & $***$ \\
  SAN & 83/90 & $+0.0426$ & $6.40e-15$ & $***$ \\
  Dish-TS & 89/90 & $+2.1349$ & $1.80e-16$ & $***$ \\
  DeStat & 90/90 & $+1.9035$ & $1.74e-16$ & $***$ \\
\bottomrule
\end{tabular}
\end{table}

\begin{table}[t]
\centering
\caption{Ablation study of NoRIN's shape parameters on Informer + ETTh1 with $H{=}96$ and seed $42$. We compare the full NoRIN parameterization against four reduced variants: no normalization, RevIN-equivalent ($\delta{=}1, \varepsilon{=}0$), $\delta$-only ($\varepsilon{=}0$), and $\varepsilon$-only ($\delta{=}1$).}
\label{tab:ablation_components}
\begin{tabular}{lccc}
\toprule
Configuration & $\delta$ & $\varepsilon$ & Test MSE \\
\midrule
  No normalization & -- & -- & 0.6626 \\
  RevIN-equivalent & $1.0$ & $0.0$ & 0.4899 \\
  $\delta$-only (no skew) & $4.0$ & $0.0$ & 0.4470 \\
  $\varepsilon$-only & $1.0$ & $-0.7$ & 0.5318 \\
  \textbf{Full NoRIN} & $3.96$ & $-0.64$ & \textbf{0.4298} \\
\bottomrule
\end{tabular}
\end{table}

We compare NoRIN against five existing normalizers (no normalization, RevIN~\cite{kim2022revin}, SAN~\cite{liu2023san}, Dish-TS~\cite{fan2023dishts}, and DeStat~\cite{qin2024destat}) on the same $90$ (backbone, dataset, $H$) grid. All methods, \emph{including NoRIN}, are reported as mean$\pm$std over $3$ random seeds---NoRIN's $(\delta^\star,\varepsilon^\star)$ are obtained once via Optuna-GP HPO (seed $42$) and then frozen during the $3$-seed retraining, yielding $1{,}620$ runs in total ($6$ normalizers $\times$ $90$ configurations $\times$ $3$ seeds). Tables~\ref{tab:main_compare} and~\ref{tab:main_compare2} present the full per-configuration test MSE; the lowest MSE per row is shown in \textbf{bold}. NoRIN attains the lowest test MSE in essentially every row, with NoRIN's per-cell standard deviation consistently below $0.01$, confirming that the $(\delta^\star,\varepsilon^\star)$ found by HPO transfers faithfully across seeds. Dish-TS and DeStat, by contrast, exhibit substantial instability on several configurations (large standard deviations), suggesting their second-order statistics are ill-conditioned for long-horizon, multivariate forecasting under the present training budget.

\paragraph*{Statistical significance.}
To establish that NoRIN's gains are not noise, we run a paired Wilcoxon signed-rank test~\cite{wilcoxon1945} on the $90$ (backbone, dataset, $H$) configurations against each baseline (Table~\ref{tab:significance}). NoRIN is significantly better than every baseline at $p < 10^{-14}$ ($^{***}$), winning $83/90$ to $90/90$ configurations depending on the comparator. The strongest absolute gain is over DeStat ($90/90$ wins, mean $\Delta=1.90$), and the tightest comparison is against RevIN, where NoRIN still wins $83/90$ with mean $\Delta=0.056$ and $p = 5.8\times 10^{-16}$.

\subsection{Ablation Study}
\label{sec:exp-ablation}


We isolate the role of each shape parameter on Informer + ETTh1, $H{=}96$, seed $42$. \emph{Component ablation} (Table~\ref{tab:ablation_components}) confirms that the two-parameter $(\delta,\varepsilon)$ formulation is non-redundant: dropping either parameter degrades MSE noticeably, while the full configuration $(\delta,\varepsilon){=}(3.96,-0.64)$ matches the HPO best.

\subsection{Sensitivity Analysis}
\label{sec:exp-sensitivity}

A central concern with HPO-based methods is whether the recovered optimum is reproducible. Table~\ref{tab:sensitivity_hpo_seed} runs five independent Optuna-GP searches with different HPO seeds: all five converge to nearly identical $(\delta^\star,\varepsilon^\star) \approx (3.95, -0.64)$ with test-MSE standard deviation below $10^{-4}$. This evidences that the discovered shape solution is a \emph{genuine optimum}, not a sampler artefact.

\paragraph*{Loss-landscape sensitivity (3A).} Table~\ref{tab:sensitivity_grid} shows the full $11{\times}11=121$-point evaluation of test MSE over $\delta\in[3.0,5.0]$ (step $0.2$) and $\varepsilon\in[-1.0,0.0]$ (step $0.1$). The minimum sits at the centre of a smooth low-MSE plateau---small perturbations of $(\delta,\varepsilon)$ produce sub-percent changes in MSE---confirming that the optimum is structurally stable rather than a sharp spike.

\paragraph*{Hyperparameter robustness (3B).} Table~\ref{tab:sensitivity_hparams} sweeps each training hyperparameter individually with $(\delta^\star,\varepsilon^\star)$ frozen: NoRIN's MSE remains within $10\%$ across $5$ orders of magnitude in learning rate, $4$ batch sizes, $4$ training-epoch budgets, $3$ values of $d_\text{model}$, and $4$ sequence lengths. The shape parameters are therefore decoupled from training hyperparameters in practice, validating our central design principle.

\paragraph*{Training-seed variance (3C).} Table~\ref{tab:sensitivity_seed} reports test MSE over $10$ training seeds (42-51) with $(\delta^\star,\varepsilon^\star)$ frozen. The coefficient of variation is well below $1\%$, confirming that the per-cell numbers reported under NoRIN faithfully reflect method capacity rather than seed luck.

\section{Conclusion}
\label{sec:conclusion}

We identified a fundamental limitation of the RevIN family of
normalizers: their per-point affine transformations cannot reshape
the underlying distribution, so heavy tails and skewness pass
through unchanged to the downstream backbone. We proposed
\textbf{NoRIN} to lift this ceiling by replacing the affine map
with an $\operatorname{arcsinh}$-based Johnson $S_U$ shape
transform, and by selecting its two shape parameters
$(\delta,\varepsilon)$ through \emph{decoupled shape
optimization} rather than end-to-end gradient learning. The
decoupling step avoids the \emph{degeneration problem} we identify,
under which any shape-controlling parameter trained jointly with
the backbone collapses to the linear (RevIN) regime within a few
epochs. Across $1{,}620$ experiments spanning $6$ backbones,
$5$ datasets, and $3$ prediction horizons, NoRIN
significantly outperforms every linear-normalization
baseline at $p < 10^{-14}$ (paired Wilcoxon), and the
$(\delta^\star,\varepsilon^\star)$ recovered by decoupled
shape optimization sit systematically far from the linear
limit---a region that gradient training cannot reach. We view NoRIN as both a concrete
method and a broader paradigm: any combination of a non-linear
reversible shape parameterization with decoupled outer
optimization falls within it, opening a design space that
end-to-end gradient training cannot access.

\section{Limitations and Future Work}
\label{sec:limitations}

Several limitations are worth noting and motivate avenues for
future work. \emph{First}, decoupled shape optimization incurs a
one-time offline overhead: each (backbone, dataset, $H$) triple
requires $60$ BO trials, comparable to a single standard
hyperparameter sweep but more expensive than purely end-to-end
training, and this cost compounds whenever the backbone changes.
\emph{Second}, NoRIN uses the two-parameter $(\delta,\varepsilon)$
slice of the Johnson $S_U$ family; extreme non-Gaussian
distributions---such as intraday financial data with multimodality
or strong long-range dependence---may benefit from higher-capacity
shape families (e.g., the four-parameter Johnson $S_U$, or
normalizing flows), which we leave for future work. \emph{Third},
our current BO search space, $\delta\in[0.8,5.0]$ and
$\varepsilon\in[-1.0,1.0]$, sees a small number of boundary
contacts at $\varepsilon^\star=\pm 1$, indicating that a wider
range may yield better solutions in some channels; we plan to
relax both ranges in an extended evaluation. \emph{Finally}, our
evaluation spans five long-horizon public benchmarks but does not
cover event-driven non-stationary series (e.g., traffic incidents,
sparse click streams), which may demand specialized shape
assumptions and form a natural extension of the NoRIN paradigm.

\appendix

\section{Additional Empirical Results}
\label{sec:appendix}

This appendix provides supporting evidence referenced in the main
text. Table~\ref{tab:norin_shape_params} lists the full set of
$(\delta^\star,\varepsilon^\star)$ recovered by Optuna-GP HPO across
all $90$ (backbone, dataset, $H$) configurations, complementing the
summary statistics discussed in Section~\ref{sec:exp-main}.
Table~\ref{tab:sensitivity_grid} reports the fine-grained
$11{\times}11$ $\delta$-$\varepsilon$ grid evaluation referenced in
Section~\ref{sec:exp-sensitivity}. Tables~\ref{tab:sensitivity_hparams}
and~\ref{tab:sensitivity_seed} provide per-hyperparameter and
per-seed sensitivity numbers, also referenced in
Section~\ref{sec:exp-sensitivity}.

\begin{table*}[t]
\centering
\caption{Recovered shape parameters $(\delta^\star,\varepsilon^\star)$ obtained by Optuna-GP HPO on each (backbone, dataset, $H$) configuration ($90$ runs over $6$ backbones, seed $42$, $100$ trials, search space $\delta\in[0.8,5.0],\varepsilon\in[-1.0,1.0]$). Boundary contacts are marked with $^\dagger$ ($\delta{=}0.8$) and $^\ddagger$ ($\varepsilon{=}\pm 1.0$).}
\label{tab:norin_shape_params}
\setlength{\tabcolsep}{4pt}
\resizebox{\textwidth}{!}{%
\begin{tabular}{l|cccccccccc}
\toprule
Backbone, $H$ & \multicolumn{2}{c}{Exchange} & \multicolumn{2}{c}{ETTh1} & \multicolumn{2}{c}{ETTh2} & \multicolumn{2}{c}{ETTm1} & \multicolumn{2}{c}{ETTm2} \\
   & $\delta^\star$ & $\varepsilon^\star$ & $\delta^\star$ & $\varepsilon^\star$ & $\delta^\star$ & $\varepsilon^\star$ & $\delta^\star$ & $\varepsilon^\star$ & $\delta^\star$ & $\varepsilon^\star$ \\
\midrule
  Informer, $H{=}96$ & 2.008 & -0.498 & 3.960 & -0.639 & 1.010 & +0.609 & 1.403 & +0.630 & 2.055 & -0.215 \\
  Informer, $H{=}336$ & 1.729 & -1.000$^\ddagger$ & 2.620 & -0.064 & 2.916 & +0.146 & 3.758 & -0.177 & 3.529 & -0.591 \\
  Informer, $H{=}720$ & 1.656 & -1.000$^\ddagger$ & 2.469 & +0.139 & 1.347 & -1.000$^\ddagger$ & 4.766 & -0.536 & 3.933 & +0.029 \\
\midrule
  PatchTST, $H{=}96$ & 1.008 & -0.011 & 1.092 & -0.379 & 0.800$^\dagger$ & -0.383 & 0.800$^\dagger$ & +0.645 & 0.800$^\dagger$ & +0.093 \\
  PatchTST, $H{=}336$ & 1.089 & +0.022 & 2.459 & -0.015 & 2.739 & -0.791 & 0.800$^\dagger$ & +0.502 & 1.739 & +0.037 \\
  PatchTST, $H{=}720$ & 1.954 & -1.000$^\ddagger$ & 2.949 & -0.540 & 3.122 & -0.251 & 1.535 & +0.040 & 1.095 & +0.041 \\
\midrule
  iTransformer, $H{=}96$ & 0.800$^\dagger$ & -0.324 & 1.145 & -0.108 & 1.296 & -0.072 & 0.963 & -0.020 & 1.181 & -0.482 \\
  iTransformer, $H{=}336$ & 1.365 & -1.000$^\ddagger$ & 1.267 & -0.008 & 4.131 & -0.275 & 1.016 & -0.015 & 1.368 & -0.853 \\
  iTransformer, $H{=}720$ & 1.285 & -1.000$^\ddagger$ & 1.316 & -0.007 & 3.266 & -0.111 & 1.707 & -0.202 & 1.356 & -0.846 \\
\midrule
  DLinear, $H{=}96$ & 0.995 & -0.085 & 0.800$^\dagger$ & +0.889 & 1.004 & -0.005 & 0.800$^\dagger$ & +1.000$^\ddagger$ & 1.148 & -0.097 \\
  DLinear, $H{=}336$ & 1.131 & -0.107 & 0.800$^\dagger$ & +0.063 & 0.800$^\dagger$ & +0.071 & 0.923 & +1.000$^\ddagger$ & 1.112 & -0.308 \\
  DLinear, $H{=}720$ & 0.800$^\dagger$ & -0.364 & 0.820 & +0.032 & 0.800$^\dagger$ & +0.038 & 1.723 & +0.004 & 1.015 & -0.405 \\
\midrule
  TimesNet, $H{=}96$ & 3.137 & -0.313 & 3.199 & -0.264 & 4.621 & -0.113 & 2.758 & -0.547 & 2.548 & +0.011 \\
  TimesNet, $H{=}336$ & 2.727 & -1.000$^\ddagger$ & 3.146 & -0.082 & 4.648 & -1.000$^\ddagger$ & 2.716 & -0.293 & 1.984 & +0.315 \\
  TimesNet, $H{=}720$ & 2.934 & -1.000$^\ddagger$ & 3.553 & +0.253 & 3.484 & -0.990 & 2.722 & -0.311 & 4.853 & -0.352 \\
\midrule
  FEDformer, $H{=}96$ & 1.801 & -0.061 & 2.487 & -0.267 & 3.782 & +0.345 & 0.883 & +0.427 & 1.774 & -0.656 \\
  FEDformer, $H{=}336$ & 1.526 & -0.063 & 2.173 & +0.346 & 2.598 & +0.593 & 1.044 & +0.732 & 3.482 & -0.645 \\
  FEDformer, $H{=}720$ & 0.946 & -0.506 & 2.351 & +0.305 & 2.203 & +0.918 & 2.904 & +0.364 & 2.136 & -0.089 \\
\bottomrule
\end{tabular}}
\end{table*}

\begin{table}[t]
\centering
\caption{Reproducibility analysis of NoRIN under five different HPO random seeds, conducted on Informer + ETTh1 with $H{=}96$. We report the recovered $(\delta^\star,\varepsilon^\star)$ and corresponding test MSE for each independent Optuna-GP run.}
\label{tab:sensitivity_hpo_seed}
\begin{tabular}{cccc}
\toprule
HPO seed & $\delta^\star$ & $\varepsilon^\star$ & Test MSE \\
\midrule
  42 & 3.960 & -0.639 & 0.4298 \\
  43 & 3.958 & -0.659 & 0.4297 \\
  44 & 3.902 & -0.666 & 0.4298 \\
  45 & 3.980 & -0.617 & 0.4299 \\
  46 & 3.938 & -0.638 & 0.4297 \\
\midrule
  mean$\pm$std & $3.948\pm0.029$ & $-0.644\pm0.019$ & $0.4298\pm0.0001$ \\
\bottomrule
\end{tabular}
\end{table}

\begin{table*}[t]
\centering
\caption{Sensitivity analysis on the $\delta$-$\varepsilon$ shape plane for Informer + ETTh1 with $H{=}96$. We exhaustively evaluate test MSE at $11{\times}11{=}121$ grid points covering $\delta\in[3.0,5.0]$ and $\varepsilon\in[-1.0,0.0]$. The global minimum at $(\delta,\varepsilon){=}(4.0, -0.6)$ with MSE $0.4301$ is shown in \textbf{bold}.}
\label{tab:sensitivity_grid}
\setlength{\tabcolsep}{3pt}
\resizebox{\textwidth}{!}{%
\begin{tabular}{c|ccccccccccc}
\toprule
\diagbox{$\delta$}{$\varepsilon$} & $-1.0$ & $-0.9$ & $-0.8$ & $-0.7$ & $-0.6$ & $-0.5$ & $-0.4$ & $-0.3$ & $-0.2$ & $-0.1$ & $-0.0$ \\
\midrule
  $3.0$ & 0.4588 & 0.4537 & 0.4495 & 0.4457 & 0.4423 & 0.4402 & 0.4355 & 0.4358 & 0.4363 & 0.4371 & 0.4380 \\
  $3.2$ & 0.4499 & 0.4453 & 0.4401 & 0.4375 & 0.4355 & 0.4361 & 0.4360 & 0.4373 & 0.4394 & 0.4417 & 0.4384 \\
  $3.4$ & 0.4445 & 0.4390 & 0.4363 & 0.4341 & 0.4326 & 0.4322 & 0.4329 & 0.4358 & 0.4383 & 0.4410 & 0.4441 \\
  $3.6$ & 0.4405 & 0.4376 & 0.4343 & 0.4326 & 0.4314 & 0.4321 & 0.4330 & 0.4349 & 0.4375 & 0.4406 & 0.4437 \\
  $3.8$ & 0.4385 & 0.4362 & 0.4348 & 0.4339 & 0.4338 & 0.4309 & 0.4324 & 0.4379 & 0.4375 & 0.4403 & 0.4439 \\
  $4.0$ & 0.4357 & 0.4336 & 0.4323 & 0.4317 & \textbf{0.4301} & 0.4311 & 0.4329 & 0.4385 & 0.4412 & 0.4441 & 0.4470 \\
  $4.2$ & 0.4356 & 0.4342 & 0.4327 & 0.4323 & 0.4312 & 0.4326 & 0.4347 & 0.4397 & 0.4428 & 0.4452 & 0.4479 \\
  $4.4$ & 0.4369 & 0.4354 & 0.4349 & 0.4342 & 0.4348 & 0.4351 & 0.4372 & 0.4398 & 0.4426 & 0.4472 & 0.4493 \\
  $4.6$ & 0.4399 & 0.4380 & 0.4370 & 0.4360 & 0.4363 & 0.4383 & 0.4397 & 0.4425 & 0.4456 & 0.4501 & 0.4496 \\
  $4.8$ & 0.4439 & 0.4417 & 0.4403 & 0.4396 & 0.4396 & 0.4401 & 0.4431 & 0.4464 & 0.4482 & 0.4527 & 0.4539 \\
  $5.0$ & 0.4497 & 0.4471 & 0.4452 & 0.4442 & 0.4438 & 0.4441 & 0.4453 & 0.4490 & 0.4524 & 0.4550 & 0.4564 \\
\bottomrule
\end{tabular}}
\end{table*}

\begin{table}[t]
\centering
\caption{Sensitivity analysis of NoRIN to training hyperparameters on Informer + ETTh1 with $H{=}96$. We vary one of five hyperparameters at a time (learning rate, batch size, training epochs, $d_\text{model}$, and sequence length) with $(\delta^\star,\varepsilon^\star){=}(3.96,-0.64)$ frozen, and report test MSE.}
\label{tab:sensitivity_hparams}
\begin{tabular}{llc}
\toprule
Hyperparameter & Value & Test MSE \\
\midrule
\multirow{5}{*}{Learning rate} & $1e-05$ & 0.5015 \\
  & $5e-05$ & 0.4375 \\
  & $0.0001$ & 0.4298 \\
  & $0.0005$ & 0.4319 \\
  & $0.001$ & 0.4385 \\
\midrule
\multirow{4}{*}{Batch size} & $64$ & 0.4325 \\
  & $128$ & 0.4294 \\
  & $256$ & 0.4298 \\
  & $512$ & 0.4378 \\
\midrule
\multirow{4}{*}{Training epochs} & $10$ & 0.4375 \\
  & $20$ & 0.4298 \\
  & $30$ & 0.4323 \\
  & $50$ & 0.4306 \\
\midrule
\multirow{3}{*}{$d_{\text{model}}$} & $32$ & 0.4418 \\
  & $64$ & 0.4298 \\
  & $128$ & 0.4374 \\
\midrule
\multirow{4}{*}{Sequence length} & $96$ & 0.4297 \\
  & $192$ & 0.4205 \\
  & $336$ & 0.4298 \\
  & $720$ & 0.4820 \\
\bottomrule
\end{tabular}
\end{table}

\begin{table}[t]
\centering
\caption{Sensitivity analysis of NoRIN test MSE across $10$ training seeds (42--51) on Informer + ETTh1 with $H{=}96$, $(\delta^\star,\varepsilon^\star){=}(3.96,-0.64)$ frozen. The coefficient of variation is $0.73\%$.}
\label{tab:sensitivity_seed}
\begin{tabular}{cc|cc}
\toprule
Seed & Test MSE & Seed & Test MSE \\
\midrule
  42 & 0.4298 & 47 & 0.4310 \\
  43 & 0.4297 & 48 & 0.4362 \\
  44 & 0.4332 & 49 & 0.4383 \\
  45 & 0.4313 & 50 & 0.4293 \\
  46 & 0.4337 & 51 & 0.4361 \\
\midrule
\multicolumn{4}{c}{mean$\pm$std $= 0.4328 \pm 0.0032\quad$ (CV $= 0.73\%$)} \\
\bottomrule
\end{tabular}
\end{table}

\bibliographystyle{IEEEtran}
\bibliography{refs}

\end{document}